\documentclass{article}

\usepackage{arxiv}

\usepackage[utf8]{inputenc}
\usepackage[T1]{fontenc}
\usepackage{amsmath}
\usepackage{amssymb}
\usepackage{amsthm}

\usepackage[table,xcdraw,dvipsnames]{xcolor}
\usepackage{graphicx,verbatim}
\usepackage{multirow}
\usepackage{booktabs}
\usepackage{enumitem}
\usepackage{subcaption}
\usepackage{url}
\definecolor{wacvblue}{rgb}{0.21,0.49,0.74}
\usepackage[pagebackref,breaklinks,colorlinks,allcolors=wacvblue]{hyperref}

\title{Auditing Data Leakage in Whole-Slide Image Multimodal Benchmarks}

\author{
  Wenhao Zhang \\
  University of Virginia \\
  \texttt{bdu8us@virginia.edu} \\
  \And
  Zhongliang Zhou \\
  Merck \& Co., Inc. \\
  \texttt{zhongliang.zhou@merck.com} \\
  \AND
  John Kang \\
  Merck \& Co., Inc. \\
  \texttt{jia.kang@merck.com} \\
  \And
  Sheng Li \\
  University of Virginia \\
  \texttt{vga8uf@virginia.edu} \\
}

\begin{document}

\maketitle

\begin{abstract}

Recent vision-language models (VLMs) for computational pathology report striking zero-shot performance on whole-slide image (WSI) visual question answering (VQA) benchmarks. We audit these claims and find them fundamentally compromised by data leakage at two hierarchical levels: patient-level leakage, where slides from the same case appear in both training and test folds, and institutional-level leakage, where different cases nonetheless share staining-batch and scanner signatures through a common Tissue Source Site (TSS). By tracing canonical slide, case, and TSS identifiers across major public resources, we document case-level train–test overlaps of 92.3–100\% on TCGA-derived benchmarks, together with near-complete TSS overlap. We further demonstrate that both leakage levels are linearly decodable from foundation-model feature space, that they induce a measurable accuracy gap between leaked and audit-clean cases on a published checkpoint, and that across multiple published WSI VLMs, peak reported accuracies concentrate on the most heavily contaminated benchmarks. Therefore, the current WSI VQA evaluation cannot distinguish genuine multimodal reasoning from nearest-neighbor retrieval over memorized institutional and patient-specific artifacts. Finally, we outline concrete recommendations for contamination-free evaluation. By addressing benchmark construction, provenance disclosure, and automated overlap auditing, we aim to guide future research toward verifiable claims of progress.

\end{abstract}
\section{Introduction}
\label{sec:intro}

\begin{figure*}[t]
\centering
\includegraphics[width=\linewidth]{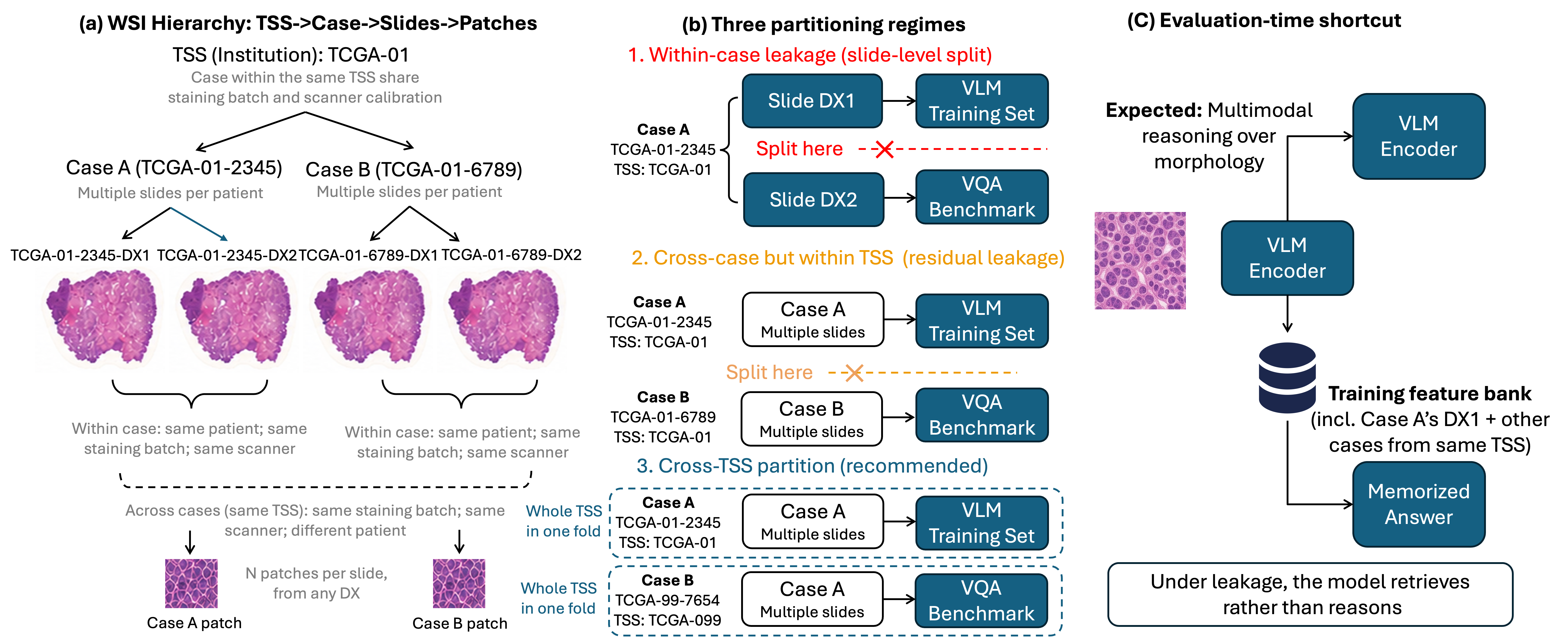}
\caption{Two-level data leakage in pathology vision-language 
models. (a) Hierarchical organization of TCGA whole-slide 
images. (b) Three partitioning regimes considered in this work. 
(c) The evaluation-time shortcut induced by either form of 
leakage.}
\label{fig:teaser}
\end{figure*}

Large labeled datasets have been critical to the success of 
vision-language models (VLMs) in computational pathology, 
enabling rapid progress on tasks such as whole-slide image (WSI) 
captioning, visual question answering (VQA), and diagnostic 
subtyping.~\cite{seyfioglu2025quiltllavavisualinstructiontuning} Recent studies routinely report state-of-the-art 
performance on standard WSI benchmarks under nominally zero-shot 
evaluation protocols, and these numbers are taken at face value 
as evidence of genuine out-of-distribution generalization.~\cite{chen2025slidechatlargevisionlanguageassistant,liang2025wsillavamultimodallargelanguage} 
However, the training datasets behind these models are 
constructed by aggressively scraping the same monolithic public  repository: The Cancer Genome Atlas (TCGA), from which the 
evaluation benchmarks are independently derived. The possibility that training and test data substantially overlap has received little systematic attention.

This overlap constitutes a fundamental methodological flaw driven by two distinct levels of leakage. At the patient level, evaluating on previously seen slides collapses zero-shot generalization into mere memorized retrieval. At the institutional level, shared Tissue Source Sites (TSS) enable models to exploit laboratory-specific batch effects (e.g., staining and scanner artifacts) as predictive shortcuts rather than learning genuine pathology. Both forms of contamination guarantee silent generalization failure when these systems are deployed in real-world, out-of-distribution clinical settings.

We present the first systematic audit of train-test 
contamination across the public WSI vision-language ecosystem. 
By tracing canonical TCGA case and slide barcodes across  
 instruction-tuning corpora and  public evaluation benchmarks , we construct a cross-contamination matrix 
that quantifies overlap at the slide, case, and institutional 
levels. The results are striking: train-test overlap on 
TCGA-derived benchmarks reaches $92.3\%$ to $100\%$ at the case 
level, with near-complete institutional (Tissue Source Site) 
overlap as well. WSI-VQA-test, one of the most widely cited 
evaluation sets, shares $100\%$ of its cases with 
WSI-Bench-train. Figure~\ref{fig:teaser} illustrates the 
hierarchical structure that gives rise to this contamination 
and the three partitioning regimes we consider.

In summary, our contributions are:

\begin{enumerate}
\item A two-level taxonomy of data leakage specific to WSI 
vision-language modeling, separating patient-level from 
institutional (TSS-level) contamination, together with a 
reproducible audit pipeline operating on canonical TCGA case, 
slide, and TSS identifiers.

\item Empirical evidence that both forms of leakage are 
linearly encoded in the foundation-model feature space, that they 
produce a measurable in-benchmark accuracy gap on a published 
checkpoint, and that this pattern reproduces in a cross-paper 
meta-analysis spanning multiple published WSI VLMs.

\item Methodological guidelines, strict slide-, patient-, and 
TSS-level deduplication, mandatory disclosure of training-test 
samples overlap at these levels, and the adoption of 
independently curated multi-center holdout sets as 
preconditions for credible progress claims in WSI VQA.
\end{enumerate}
\section{Related Work}
\label{sec:related_work}

\subsection{WSI Multimodal Benchmarks and Models}
\label{sec:rw_ecosystem}

The pathology multimodal literature comprises a tightly coupled set of
training corpora, benchmarks, and models. Two structural facts about
this ecosystem motivate our audit.

First, slide-level supervision has grown by roughly two orders of
magnitude within two years. Early efforts such as
ARCH~\cite{gamper2021multiple} mined $7.6$K image-caption groups from
textbooks. Web-scale patch corpora including
OpenPath~\cite{huang2023plip}, Quilt-1M~\cite{ikezogwo2023quilt1m},
and CONCH~\cite{Lu_2023_CVPR} then emerged, followed by
PathGen-1.6M~\cite{sun2025pathgen} and slide-level instruction-tuning
resources such as WSI-VQA~\cite{chen2024wsivqa},
SlideInstruction~\cite{chen2025slidechat},
WSI-Bench~\cite{liang2025wsillavamultimodallargelanguage}, and TITAN~\cite{ding2025titan},
together with the report-generation corpora
WsiCaption~\cite{chen2024wsicaptionmultipleinstancegeneration} and
HistGen~\cite{guo2024histgen}. Across this trajectory, TCGA dominates
the source pool, with BCNB and CPTAC the only widely used
alternatives, since the joint requirement of diagnostic-grade WSIs,
free redistribution, and paired reports admits no comparable source.
The resulting structural convergence, in which many corpora draw from
the same case pool and are then re-released as both training and
evaluation splits, is the precondition for the contamination we
quantify in Section~\ref{sec:overlap_audit}. We exclude patch-level
corpora such as OpenPath and Quilt-1M from the audit because their
tiles carry no canonical patient or slide identifier and are not the
substrate of slide-level VQA.

Second, the VLMs that consume these corpora are evaluated almost
exclusively on the benchmarks those same corpora spawn. Branching
from LLaVA-Med~\cite{li2024llavamed}, pathology-specific systems span
three design axes. Along the input-scale axis, patch-level systems
such as PathAsst~\cite{zhang2025pathor1multimodalreinforcementlearningbased},
Quilt-LLaVA~\cite{seyfioglu2024quiltllava}, and
PathChat~\cite{lu2024multimodal} coexist with slide-level systems
including SlideChat~\cite{chen2025slidechat},
WSI-LLaVA~\cite{liang2025wsillavamultimodallargelanguage}, and
TITAN~\cite{ding2025titan}, the last pretrained natively over
$8{,}192 \times 8{,}192$ regions. Along the unification axis,
CPath-Omni~\cite{sun2024cpathomni} consolidates patch- and slide-level
tasks within a single $15$B-parameter LMM. Along the reasoning and
agency axis, Patho-R1~\cite{zhang2025pathor1multimodalreinforcementlearningbased} and
PathReasoner-R1~\cite{jiang2026pathreasoner} adopt RL-style
post-training, while CPathAgent~\cite{sun2025cpathagent} and
PathAgent~\cite{chen2025pathagent} emulate the pathologist's
zoom-and-pan workflow. Every system reports headline state-of-the-art
numbers on TCGA-derived test sets, and none report training-test
overlap at the case or TSS level. As
Section~\ref{sec:meta_analysis} shows, the most contaminated
benchmarks host the strongest reported gains, leaving it unclear
whether current accuracy reflects multimodal reasoning or retrieval
over memorized reports.

\subsection{Data Leakage in Medical Imaging and Digital Pathology}
\label{sec:rw_leakage}

Leakage is a documented shortcut hazard across medical imaging.
DeGrave et al.~\cite{degrave2021ai} showed that COVID-19
chest-radiograph models latched onto hospital-specific markers rather
than pulmonary findings.
Roberts et al.~\cite{roberts2021common} reported that none of $415$
reviewed studies were of sufficient quality for clinical use, citing
patient duplication and inadequate external validation among the most
common failure modes.

Within pathology, prior findings form a hierarchy of granularity that
our audit directly extends. At the patch-within-slide level,
Bussola et al.~\cite{bussola2021ai} demonstrated up to $41$~pp
inflation when tiles from a single patient appear in both training
and validation folds. At the slide-within-patient level,
Oner et al.~\cite{oner2020training} reanalyzed a prominent
TCGA-LUAD model and showed that its apparent performance collapsed
once patient-level splitting was enforced. At the
patient-within-institution level,
Howard et al.~\cite{howard2021impact} established that TSS identity
is recoverable from H\&E features with AUROC up to $0.998$ and
proposed preserved-site cross-validation as a remedy. Subsequent work
by de Jong et al.~\cite{de2025current} showed that the same TSS signature
persists in current pathology foundation models, indicating that the
hazard has scaled with rather than been mitigated by modern encoders.

These prior audits share two limitations that our work addresses.
First, all are confined to unimodal CNNs or self-supervised foundation
models, and to classification or representation probing; none audit a
multimodal VLM corpus. Second, each leakage level has been studied in
isolation. We provide the first joint audit of patient- and
institutional-level leakage across the multimodal WSI VLM ecosystem.
In this setting the leakage surface is broader, since each
contaminated case carries an entire instruction-tuning record rather
than a single label, and the downstream consequence is more severe,
since a contaminated benchmark cannot distinguish multimodal reasoning
from textual retrieval of memorized reports.
\section{Formalizing Data Leakage in Pathology VLMs}
\label{sec:data_leakage}

Unlike natural image benchmarks, where data leakage
typically manifests as exact image duplication,
whole-slide images (WSIs) admit a hierarchical form of
contamination. This section first describes the
architectural paradigm of WSI VLMs that motivates
auditing at the encoder feature level, then formalizes
the hierarchical contamination structure, defines two
levels of leakage, and specifies the metrics used in our
audit.

\subsection{The Offline-Encoder Paradigm of WSI VLMs}
\label{sec:paradigm}

A defining characteristic of the WSI VLM ecosystem is
that visual representation learning and instruction
tuning are decoupled into two strictly sequential stages,
following the LLaVA-style
recipe~\cite{liu2023visualinstructiontuning, li2024llavamed}. In Stage 1,
each WSI is processed once through a pretrained visual
encoder to produce a fixed slide-level or patch-bag
representation. In Stage 2,
a lightweight projector and a large language model are
trained on these cached features paired with
instruction-following text. The visual encoder itself
remains frozen throughout instruction tuning.

Within this universal paradigm, current WSI VLMs differ
in how their frozen encoder is obtained, and we
distinguish two categories. The first one adopts an
off-the-shelf pathology foundation model without further
modification. Representative examples include
SlideChat~\cite{chen2025slidechat}, which composes the
patch-level encoder CONCH~\cite{Lu_2023_CVPR} with a
LongNet-based slide-level
encoder~\cite{ding2023longnet}, and
WSI-LLaVA~\cite{liang2025wsillavamultimodallargelanguage}, which uses a
dual-scale patch and slide encoder built on pretrained
components. The second one pretrains a pathology-domain
CLIP variant specifically for the VLM. Representative
examples include Quilt-LLaVA~\cite{seyfioglu2024quiltllava}, which uses
Quilt-CLIP pretrained on
Quilt-1M~\cite{ikezogwo2023quilt1m}, and PathGen-LLaVA, which uses
PathGen-CLIP-L pretrained on
PathGen-1.6M~\cite{sun2025pathgen}. The two categories
differ in the provenance and supervision of their
encoder, but they are identical in one respect that
matters for our audit: in both cases, the encoder is
fixed before instruction tuning begins and is not
updated during it.

This shared design is not a stylistic choice but a
computational necessity. A single diagnostic WSI
typically contains tens of thousands of $224 \times 224$
patches, so end-to-end backpropagation through the
encoder during instruction tuning is intractable at
scale. As a direct consequence, the visual
representation that a WSI VLM operates on is fixed by
its choice of encoder. Any patient or institutional
signal that the encoder embeds into the feature space
becomes a permanent input to the projector and the LLM.
The downstream parameters can only reweight what the
encoder has already encoded; they cannot remove a
confounder that is already present. This observation
motivates the auditing strategy adopted in
Section~\ref{sec:feature_evidence}, where we probe
representative encoders from both categories. Probing the frozen encoder's feature space is methodologically
equivalent to probing the visual input of every VLM
built on that encoder, while avoiding the confounds
introduced by downstream projector and LLM hidden
states.

\subsection{Hierarchical Structure and Two Levels of
Leakage}
\label{sec:two_levels}

We begin with the WSI data hierarchy. Every diagnostic
WSI in the TCGA repository is identified by a barcode
encoding a three-level nesting~\cite{saltz2018til}:
\begin{equation}
\underbrace{\texttt{TCGA-BH}}_{\text{TSS}}\texttt{-}
\underbrace{\texttt{A0BS}}_{\text{Case}}\texttt{-}
\underbrace{\texttt{01Z-00-DX1}}_{\text{Slide}}
\;\longrightarrow\;
\{\,p_1, p_2, \ldots, p_N\,\}_{\text{Patches}}.
\label{eq:barcode}
\end{equation}
First, the Tissue Source Site (TSS) identifies the
contributing institution. Second, the case identifier
maps one-to-one to a patient. Third, the slide suffix
(DX1, DX2, $\ldots$) distinguishes tissue sections taken
from the same case. We write $\mathrm{case}(s)$ for the
case to which slide $s$ belongs and $\mathrm{tss}(c)$ for the TSS to which case $c$ belongs. Sibling slides from a
single case share patient phenotype, staining batch, and
scanner calibration, while cases from a single TSS share
laboratory-level staining and scanning signatures.

Given a training dataset $\mathcal{D}_{\mathrm{train}}$
and an evaluation benchmark
$\mathcal{D}_{\mathrm{test}}$, each consisting of a set
of slides, we define two contamination conditions. First,
a test slide $s \in \mathcal{D}_{\mathrm{test}}$ is
patient-leaked if there exists a training slide
$s' \in \mathcal{D}_{\mathrm{train}}$ drawn from the same
case:
\begin{equation}
\begin{split}
\mathcal{L}_{\mathrm{patient}} & = 
\bigl\{\, s \in \mathcal{D}_{\mathrm{test}} \mid
\exists\, s' \in \mathcal{D}_{\mathrm{train}} : \\
& \qquad \mathrm{case}(s) = \mathrm{case}(s') \,\bigr\}.
\end{split}
\label{eq:patient_leak}
\end{equation}
Second, a test slide is institution-leaked if it
is not patient-leaked but shares a TSS with some training
slide:
\begin{equation}
\begin{split}
\mathcal{L}_{\mathrm{TSS}} & = 
\bigl\{\, s \in \mathcal{D}_{\mathrm{test}} \setminus \mathcal{L}_{\mathrm{patient}} \mid
\exists\, s' \in \mathcal{D}_{\mathrm{train}} : \\
& \qquad \mathrm{tss}\bigl(\mathrm{case}(s)\bigr) = \mathrm{tss}\bigl(\mathrm{case}(s')\bigr) \,\bigr\}.
\end{split}
\label{eq:tss_leak}
\end{equation}
A test slide belonging to neither set is
audit-clean. The three sets partition the test
benchmark exhaustively,
\begin{equation}
\mathcal{D}_{\mathrm{test}} \;=\;
\mathcal{L}_{\mathrm{patient}} \;\cup\;
\mathcal{L}_{\mathrm{TSS}} \;\cup\;
\mathcal{D}_{\mathrm{clean}},
\qquad
\label{eq:partition}
\end{equation}
and a benchmark qualifies as a valid zero-shot evaluation
only when $\mathcal{D}_{\mathrm{clean}} =
\mathcal{D}_{\mathrm{test}}$. To quantify contamination,
we report the case-level and TSS-level overlap ratios,
\begin{equation}
\rho_{\mathrm{case}} \;=\;
\frac{|\,\mathcal{C}_{\mathrm{test}} \cap
\mathcal{C}_{\mathrm{train}}\,|}
{|\,\mathcal{C}_{\mathrm{test}}\,|}
\;, \\
\rho_{\mathrm{TSS}} \;=\;
\frac{|\,\mathcal{T}_{\mathrm{test}} \cap
\mathcal{T}_{\mathrm{train}}\,|}
{|\,\mathcal{T}_{\mathrm{test}}\,|},
\label{eq:overlap}
\end{equation}
where $\mathcal{C}$ and $\mathcal{T}$ denote the sets of
cases and TSSs underlying the corresponding slide
collections. A benchmark with $\rho_{\mathrm{case}} = 1$
has been entirely subsumed at the patient level, whereas
$\rho_{\mathrm{TSS}} = 1$ indicates that no institutional generlizationablity remains.

Two scope clarifications are in order. First, we restrict
this audit to the instruction-tuning stage of the VLM
pipeline, where structured TCGA metadata renders
$\rho_{\mathrm{case}}$ and $\rho_{\mathrm{TSS}}$
computable via Eq.~\eqref{eq:overlap}. Upstream leakage
through the visual encoder's pretraining corpus is
acknowledged but falls outside our scope, since such
corpora are typically not publicly available. Second, we exclude
ROI-level and tile-level datasets such as
OpenPath~\cite{huang2023plip} and
Quilt-1M~\cite{ikezogwo2023quilt1m}, which lack canonical patient or slide identifiers and for which the functions
$\mathrm{case}(\cdot)$ and $\mathrm{tss}(\cdot)$ are
undefined. The structural reason such high overlap arises
in WSI corpora is that TCGA is the only public repository
satisfying the prerequisites for multimodal instruction
tuning, which forces all major training dataset and
benchmarks to draw from the same pool.
\section{Experiments}
\label{sec:experiments}

This section presents four lines of evidence that the
contamination formalized in Section~\ref{sec:data_leakage}
is pervasive and consequential. We first quantify
train-test overlap across the public WSI VLM ecosystem
(\S\ref{sec:overlap_audit}). We then verify that both
levels of leakage leave detectable traces in
foundation-model feature space
(\S\ref{sec:feature_evidence}), using two structurally
distinct encoders to show the effect is not an artifact
of any single representation. Finally, we demonstrate
that these traces translate into measurable benchmark
inflation, across multiple published models
(\S\ref{sec:meta_analysis}).

\subsection{Cross-Dataset Overlap Audit}
\label{sec:overlap_audit}

For every pairing of training corpus and evaluation benchmark, we compute both $\rho_{\mathrm{case}}$
 and $\rho_{\mathrm{TSS}}$ (Eq.~\ref{eq:overlap}) by matching canonical TCGA case and tissue-source-site (TSS) identifiers. Figure~\ref{fig:leakage_heatmap} reports the resulting two-panel overlap matrix spanning eight instruction-tuning corpora and four TCGA-derived test benchmarks. Our audit covers the most widely used public WSI datasets, including SlideBench~\cite{chen2025slidechatlargevisionlanguageassistant}, WSI-Bench~\cite{liang2025wsillavamultimodallargelanguage}, WSI-VQA~\cite{chen2024wsivqa}, HistGen~\cite{guo2024histgen}, and PathGen-1.6M~\cite{sun2025pathgen}. Three findings emerge:

\begin{figure*}[t]
    \centering
    \includegraphics[width=1.0\linewidth]{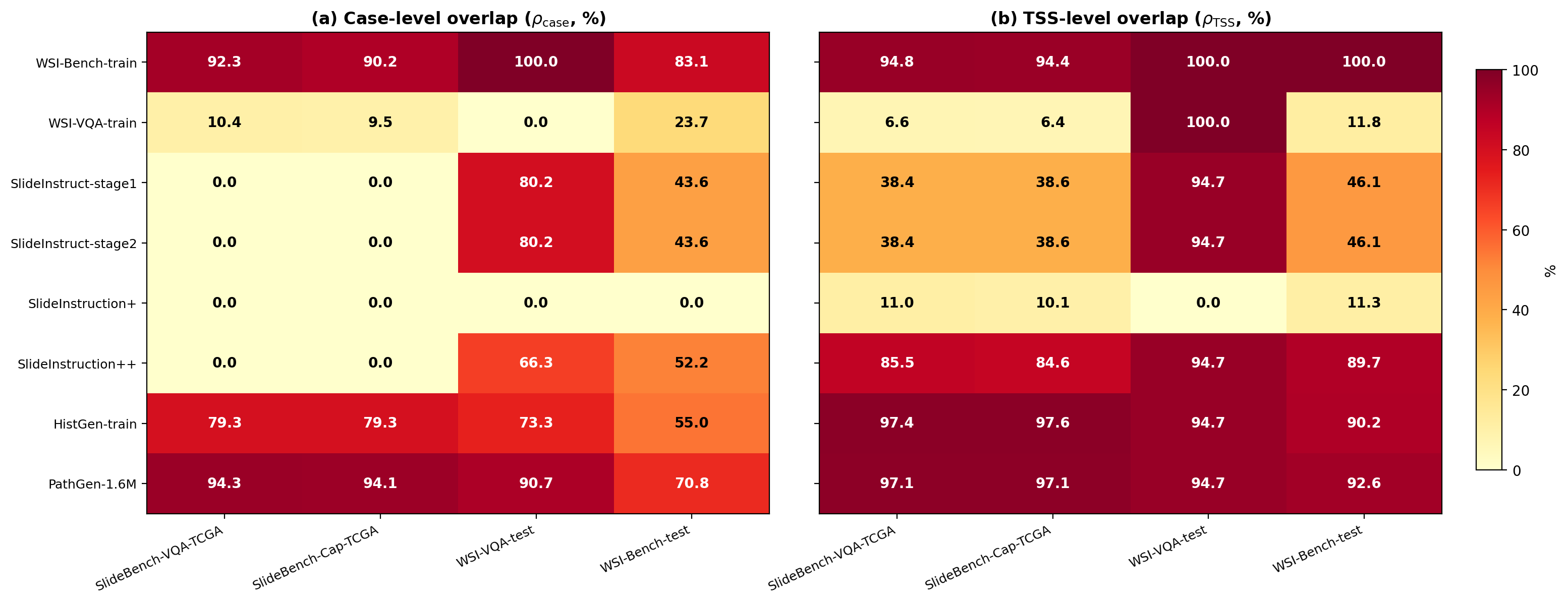}
    \caption{Two-panel train-test overlap matrix across
    pathology VLM training corpora (rows) and four
    TCGA-derived evaluation benchmarks (columns).
    (a)~Case-level overlap $\rho_{\mathrm{case}}$ (\%).
    (b)~TSS-level overlap $\rho_{\mathrm{TSS}}$ (\%).
    TCGA-derived benchmarks are heavily contaminated at
    both levels. TSS-level overlap is uniformly higher
    than case-level overlap, revealing institutional
    contamination even where case-level deduplication
    appears successful. Non-TCGA benchmarks (BCNB,
    CPTAC, HistAI) trivially maintain $0\%$ at both
    levels and are omitted.}
    \label{fig:leakage_heatmap}
\end{figure*}

First, case-level overlap on TCGA-derived benchmarks is
extreme across most training corpora. WSI-Bench-train
overlaps with $92.3\%$, $90.2\%$, $100\%$, and $83.1\%$
of SlideBench-VQA-TCGA, SlideBench-Cap-TCGA,
WSI-VQA-test, and WSI-Bench-test cases, respectively.
PathGen-1.6M shows comparable contamination at $94.3\%$,
$94.1\%$, $90.7\%$, and $70.8\%$. WSI-VQA-test is fully
subsumed by WSI-Bench-train at $\rho_{\mathrm{case}} =
100\%$, meaning every test patient was seen during
training. A particularly striking case is internal to a
single benchmark: WSI-Bench-train and WSI-Bench-test
themselves share $83.1\%$ of their cases, indicating
that the benchmark curators' own data partition does
not separate patients between the two halves.

Second, TSS-level overlap is uniformly higher than
case-level overlap and reveals institutional
contamination that a case-level audit alone would miss.
Two pairs make this particularly clear. WSI-VQA-train
and WSI-VQA-test exhibit $0\%$ case overlap, indicating
that the WSI-VQA authors correctly performed
patient-level deduplication, yet their TSS overlap is
$100\%$. Every test patient comes from an institution 
already represented in training. SlideInstruction++
overlaps with SlideBench-VQA-TCGA at $0\%$ case but
$85.5\%$ TSS, and with WSI-VQA-test at $66.3\%$ case
but $94.7\%$ TSS. This pattern shows that patient-level
deduplication, while necessary, leaves the same
staining, scanning, and laboratory batch effects in
both train and test splits, satisfying the formal
definition of $\mathcal{L}_{\mathrm{TSS}}$ leakage
(Eq.~\ref{eq:tss_leak}).

Third, attempted decontamination is achievable at the
case level but remains incomplete at the TSS level.
SlideInstruction+ is the only training corpus we
audited that achieves $\rho_{\mathrm{case}} = 0\%$
against all four TCGA-derived test benchmarks. Even so,
it retains $11.0\%$ TSS overlap with SlideBench-VQA-TCGA,
$10.1\%$ with SlideBench-Cap-TCGA, and $11.3\%$ with
WSI-Bench-test. Eliminating patient-level overlap can
be accomplished by metadata filtering, but eliminating
institutional overlap requires sourcing test data from
Tissue Source Sites not represented in any major public
training corpus, which TCGA-derived benchmarks
structurally cannot provide. This is why genuine
zero-shot evaluation in the current ecosystem is
feasible only on independently curated, non-TCGA
cohorts.

\subsection{Feature-Space Evidence of Leakage}
\label{sec:feature_evidence}

The overlap documented above is defined at the level of
metadata identifiers. A natural question is whether this
identifier-level overlap translates into measurable
redundancy in the learned feature representations that
VLMs actually consume. We address this with two
complementary experiments, a pairwise similarity analysis
and a linear probing study, and run both on two
structurally distinct slide encoders. The first is
\textbf{CONCH-v1}~\cite{Lu_2023_CVPR}, a patch-level
contrastive encoder whose slide representation we obtain
by mean-pooling $\ell_2$-normalized patch features. The
second is \textbf{TITAN}~\cite{ding2025titan}, a natively
slide-level pretrained encoder. If leakage were merely an
artifact of patch-level mean-pooling, it would not
survive in TITAN's native slide embeddings. Both
experiments are computed on the $6{,}756$ slides for
which both encoders' features are available, so the two
encoders are always compared on an identical slide
population.

\paragraph{Pairwise similarity.}
For each encoder we compute cosine similarity between
slide embeddings across three configurations of
hierarchical proximity: (i)~intra-patient pairs (sibling
slides from the same case), (ii)~intra-TSS, cross-patient
pairs (different patients from the same Tissue Source
Site), and (iii)~cross-TSS pairs (slides from different
institutions, $n{=}20{,}000$ sampled pairs). The two
encoders occupy feature spaces with different intrinsic
scales. TITAN's cross-TSS baseline similarity is $0.233$,
while CONCH-v1's is $0.664$. Absolute similarity values
are therefore not directly comparable across encoders. We
report each group relative to its own encoder's
cross-TSS baseline, isolating the excess similarity
attributable to shared patient or institutional identity.

\begin{figure}[t]
    \centering
    \includegraphics[width=0.7\linewidth]{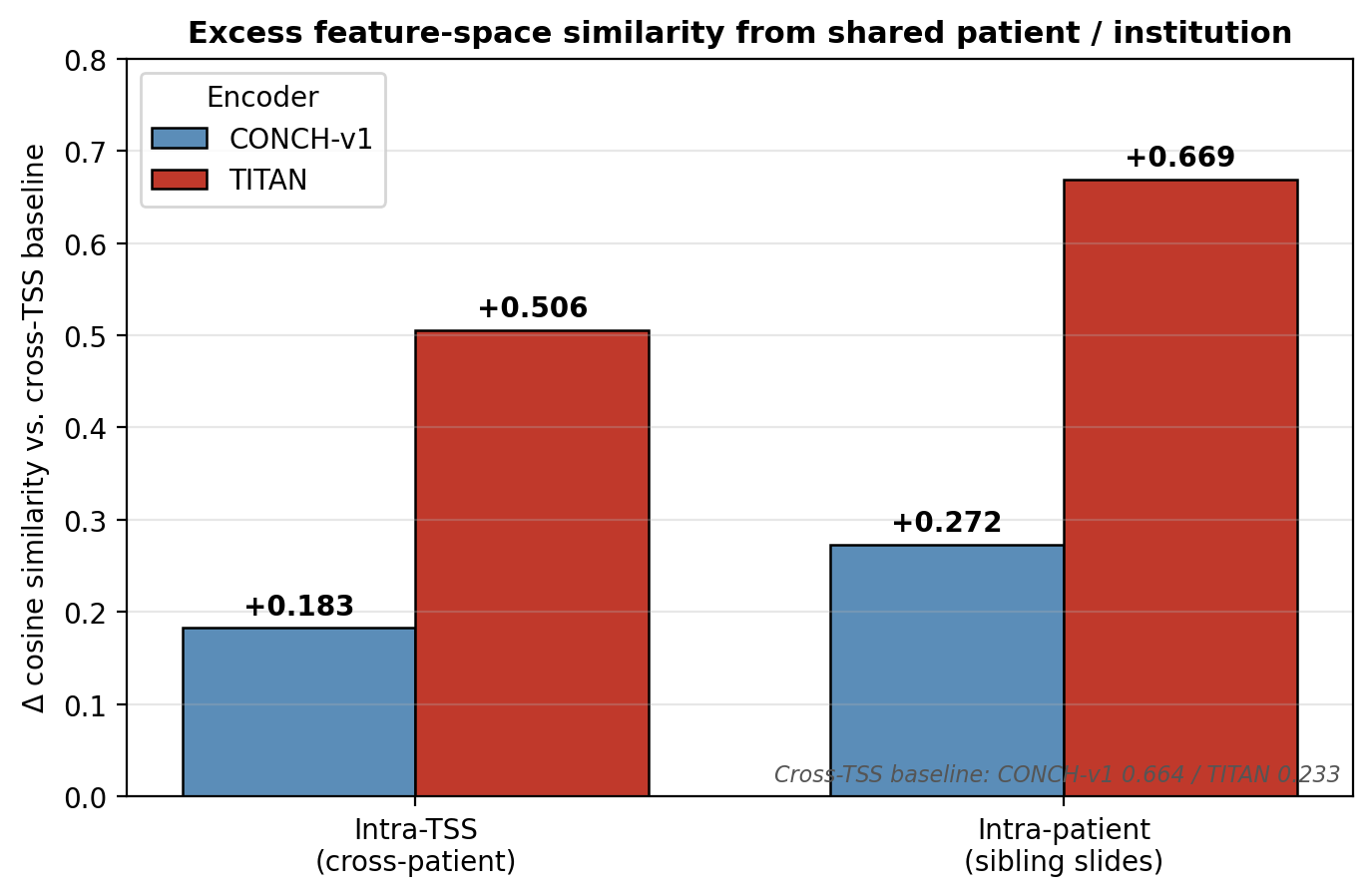}
    \caption{Excess slide-level cosine similarity over
    each encoder's own cross-TSS baseline, on the shared
    set of $6{,}756$ slides. For both encoders, similarity
    increases monotonically with hierarchical proximity
    (cross-TSS $<$ intra-TSS $<$ intra-patient), and the
    increase is markedly steeper for TITAN, the natively
    slide-level encoder. Absolute baseline similarities
    are $0.664$ (CONCH-v1) and $0.233$ (TITAN).}
    \label{fig:feature_collision}
\end{figure}

Figure~\ref{fig:feature_collision} reveals a monotonic
separation across the three groups for both encoders. For
CONCH-v1, intra-TSS pairs sit $+0.183$ above the
cross-TSS baseline and intra-patient pairs $+0.272$ above
it. For TITAN the same ordering holds and the gaps are
substantially larger, $+0.506$ for intra-TSS and $+0.669$
for intra-patient pairs. All four transitions are
statistically significant (Mann-Whitney $U$, $p<10^{-200}$). Two conclusions follow: First, the monotonic
structure (cross-TSS $<$ intra-TSS $<$ intra-patient)
reproduces across two encoders with entirely different
pretraining objectives and input granularities. This
establishes that the effect is a property of pathology
slide content rather than of any single representation.
Second, the natively slide-level encoder does not
attenuate the effect. TITAN's intra-patient excess
similarity ($+0.669$) is more than double CONCH-v1's
($+0.272$), indicating that slide-level pretraining
concentrates rather than dilutes patient-identifying
signal. These two configurations map directly onto the
leakage levels. Intra-patient proximity realizes
$\mathcal{L}_{\mathrm{patient}}$, and intra-TSS proximity
realizes $\mathcal{L}_{\mathrm{TSS}}$.

\paragraph{Linear probing for latent confounders.}
If patient and institutional signatures are encoded in
the feature space, a linear classifier should recover
them. We formulate two probing tasks on each encoder's
frozen slide embeddings, patient re-identification (ReID)
restricted to multi-slide cases, and TSS detection. For
each task we retain only classes with at least two
slides, use a stratified $70/30$ train-test split, and
fit a logistic-regression probe. ReID is scored by top-1
accuracy against the chance rate
$1/(\text{\#patients})$. TSS detection is scored by
one-versus-rest macro AUROC.

\begin{figure}[t]
    \centering
    \includegraphics[width=\linewidth]{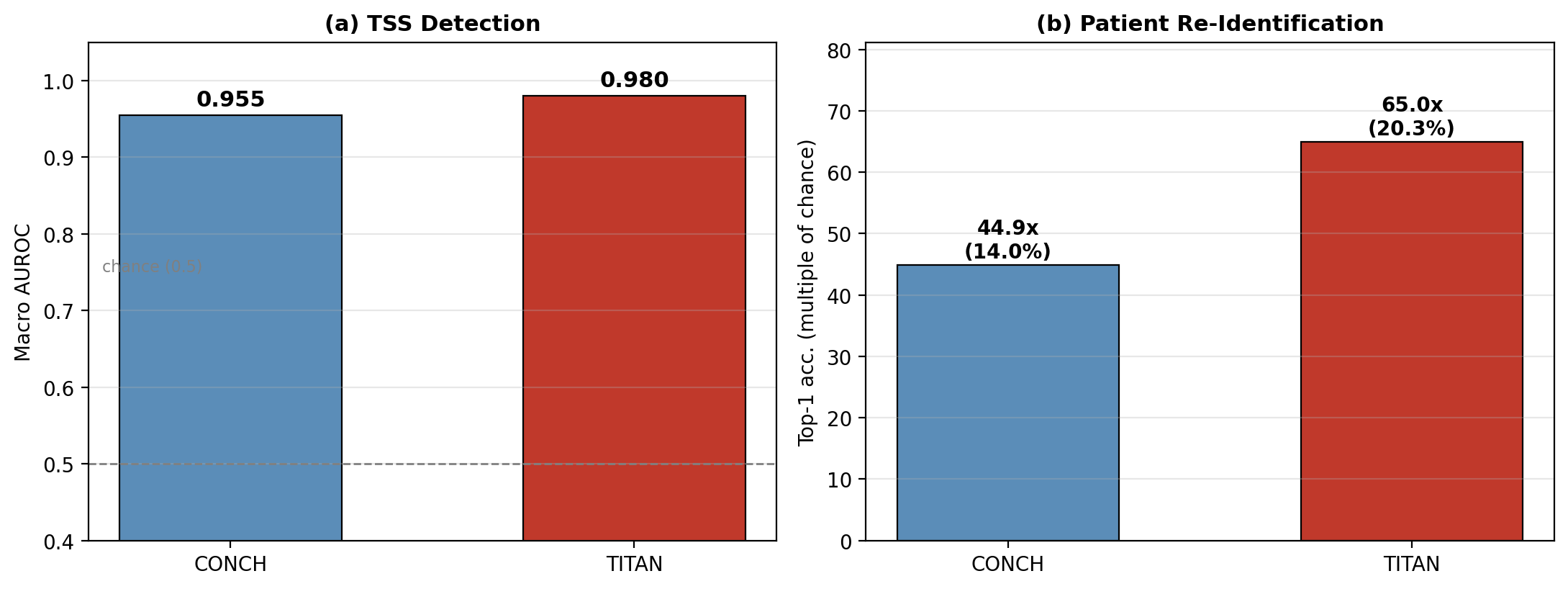}
    \caption{Linear probing on frozen slide embeddings
    for two encoders, on the shared $6{,}756$-slide
    population. (a)~TSS detection macro AUROC.
    (b)~Patient ReID top-1 accuracy as a multiple of the
    chance baseline (absolute accuracy in parentheses).
    Both confounders are recoverable far above chance for
    both encoders, and the natively slide-level encoder
    (TITAN) yields the stronger leakage signal on both
    tasks.}
    \label{fig:probing_results}
\end{figure}

Figure~\ref{fig:probing_results} confirms that both
confounders are linearly separable in both encoders.
Patient ReID reaches $14.0\%$ for CONCH-v1 and $20.3\%$
for TITAN, corresponding to $44.9\times$ and $65.0\times$
their respective chance baselines. TSS detection yields
macro AUROC of $0.955$ and $0.980$. Two points deserve
emphasis. First, Howard et al.~\cite{howard2021impact}
reported in 2021 that ResNet-50 features could detect TSS
with AUROC~$>0.95$. Four years and two generations of
pathology foundation models later, the institutional
confound has not been mitigated, but even
sharpened. Second, the consistent ordering
TITAN~$>$~CONCH-v1 on both tasks mirrors the
pairwise-similarity finding. The encoder designed
specifically for slide-level representation encodes
patient and institutional identity most explicitly. Any
downstream VLM consuming these features therefore
inherits both the patient-identity leak and the
institutional batch-effect leak. This concern is
increasingly relevant because slide-level VLMs are
increasingly adopting exactly this class of encoder.

\subsection{Cross-Paper Meta-Analysis}
\label{sec:meta_analysis}

The preceding experiments establish that overlap exists
(\S\ref{sec:overlap_audit}) and that it produces
detectable feature-space redundancy
(\S\ref{sec:feature_evidence}). A remaining question is
whether this redundancy inflates the performance numbers
reported in the published literature.

\begin{table*}[t]
\centering
\caption{Published accuracy (\%) on three WSI VQA
benchmarks alongside audit-verified case-level leakage
(\%). \colorbox{red!22}{Red} marks $\rho_{\mathrm{case}}
> 80\%$. \colorbox{orange!25}{Orange} marks $20\%$ to
$80\%$. White cells are audit-clean ($0\%$). $\diamond$:
reproduced from MLLM-HWSI
Table~3~\cite{alawode2026mllmhwsi}.
\textsuperscript{*}: SlideChat-filtered subset (filter
criterion unspecified). \textsuperscript{$\dag$}:
$80.2\%$ case overlap with unfiltered SlideInstruct.}
\label{tab:meta_analysis}
\resizebox{\textwidth}{!}{%
\begin{tabular}{@{} l l l c c c c @{}}
\toprule
\multirow{2}{*}{Model} & \multirow{2}{*}{Venue}
& \multirow{2}{*}{Training data}
& \multicolumn{2}{c}{Accuracy (\%)}
& \multicolumn{2}{c}{$\rho_{\mathrm{case}}$ (\%)} \\
\cmidrule(lr){4-5} \cmidrule(l){6-7}
& & & SB-TCGA & WSI-VQA
& SB-TCGA & WSI-VQA \\
\midrule
Quilt-LLaVA~\cite{seyfioglu2024quiltllava}
& CVPR '24 & Quilt-1M (patch)
& 40.2$\diamond$ & 35.4$\diamond$
& 0 & -- \\
WSI-VQA~\cite{chen2024wsivqa}
& ECCV '24 & WSI-VQA-train
& 27.6$\diamond$ & \cellcolor{orange!25}46.9$\diamond$
& \cellcolor{orange!25}10.4 & internal \\
SlideChat~\cite{chen2025slidechat}
& CVPR '25 & SlideInstruct
& 81.2 & \cellcolor{red!22}60.1\textsuperscript{*}
& 0 & \cellcolor{red!22}\textbf{80.2}\textsuperscript{$\dag$} \\
PathGen-LLaVA~\cite{sun2025pathgen}
& ICLR '25 & PathGen-1.6M
& \cellcolor{red!22}46.5 & \cellcolor{red!22}33.1$\diamond$
& \cellcolor{red!22}\textbf{94.3} & \cellcolor{red!22}\textbf{90.7} \\
WSI-LLaVA~\cite{liang2025wsillavamultimodallargelanguage}
& arXiv '24/12 & WSI-Bench-train
& \cellcolor{red!22}82.5$\diamond$ & \cellcolor{red!22}54.6$\diamond$
& \cellcolor{red!22}\textbf{92.3} & \cellcolor{red!22}\textbf{100.0} \\
MLLM-HWSI~\cite{alawode2026mllmhwsi}
& arXiv '26/03 & WSI-Bench+
& \cellcolor{red!22}89.6 & \cellcolor{red!22}69.2
& \cellcolor{red!22}\textbf{92.3} & \cellcolor{red!22}\textbf{100.0} \\
\bottomrule
\end{tabular}%
}
\end{table*}

Table~\ref{tab:meta_analysis} reveals two patterns.
First, the highest reported accuracies on
SlideBench-VQA-TCGA and WSI-VQA-test consistently
co-occur with severe contamination
($\rho_{\mathrm{case}} > 90\%$, red cells). MLLM-HWSI
reports $89.6\%$ on SlideBench-VQA-TCGA with $92.3\%$
case overlap. WSI-LLaVA reports $82.5\%$ with identical
overlap. Both report on WSI-VQA-test with $100\%$ case
overlap. Second, models whose training data is non-TCGA
or patch-level (LLaVA-Med, Quilt-LLaVA, TITAN) show no
detectable case-level overlap under our identifier-based
audit. We stress that this is a lower bound rather than a
certificate of cleanliness: the pretraining provenance of
these models is either undisclosed or
composed of patch-level web sources that carry no
canonical case or slide identifiers, so the
functions $\mathrm{case}(\cdot)$ and $\mathrm{tss}(\cdot)$
are undefined on their training data and Eq.~\ref{eq:overlap}
cannot resolve any overlap that may nonetheless exist. The horizontal rule
in the table separates models with no detectable case-level
contamination (above) from models with at least one
contaminated (model, benchmark) pair (below).

A notable finding concerns the community's partial
awareness of this issue.
SlideChat~\cite{chen2025slidechat} (CVPR~2025) explicitly
acknowledges overlap with WSI-VQA-test and constructs a
filtered subset (WSI-VQA*), reporting $60.1\%$ on it.
However, the filter criterion is unspecified, the
pre-filter accuracy is unreported, and no subsequent SOTA
paper replicates this filtering step. Five models
published after SlideChat, namely WSI-LLaVA,
PathGen-LLaVA, CPath-Omni and MLLM-HWSI, report
on the unfiltered, fully contaminated WSI-VQA-test as
zero-shot external validation without
acknowledgment.

These observations are consistent with the hypothesis
that contaminated benchmarks inflate reported accuracy.
However, cross-benchmark comparisons confound leakage
with intrinsic task difficulty. The next experiment
controls for this confound.

\subsection{Within-Benchmark Stratification}
\label{sec:within_benchmark}

To control for task difficulty, we exploit the fact that
SlideBench-VQA-TCGA contains both leaked and audit-clean
patients within the same test set. Our audit identifies leaked slides and audit-clean slides
among its $1{,}292$ total. We run inference using the
published WSI-LLaVA checkpoint and stratify accuracy by
leakage status across cancer type and question category.

\begin{table}[t]
\centering
\small
\caption{WSI-LLaVA accuracy on SlideBench-VQA-TCGA,
stratified by audit-verified leakage status across cancer
type and question category. Leaked accuracy exceeds clean
accuracy by $3$ to $8$ pp across all well-powered
subsets.}
\label{tab:within_cancer}
\begin{tabular}{@{}lccc@{}}
\toprule
Subset & Leaked & Clean & Gap \\
\midrule
Microscopy Qs
& $\mathbf{71.8\%}$ {\scriptsize($n{=}78$)}
& $64.3\%$ {\scriptsize($n{=}14$)}
& $+7.5$ \\
UCEC stratum
& $\mathbf{67.4\%}$ {\scriptsize($n{=}190$)}
& $61.1\%$ {\scriptsize($n{=}36$)}
& $+6.3$ \\
UCEC$+$COAD pool
& $\mathbf{65.9\%}$ {\scriptsize($n{=}325$)}
& $62.1\%$ {\scriptsize($n{=}58$)}
& $+3.8$ \\
Diagnosis Qs
& $\mathbf{64.8\%}$ {\scriptsize($n{=}233$)}
& $61.4\%$ {\scriptsize($n{=}44$)}
& $+3.4$ \\
\bottomrule
\end{tabular}
\end{table}

Table~\ref{tab:within_cancer} reports the results. Across
every well-powered subset, leaked accuracy exceeds clean
accuracy. Gaps range from $+3.4$ pp on Diagnosis
questions to $+7.5$ pp on Microscopy questions, the
category most dependent on fine-grained visual
recognition. The direction is positive in all four
independent subsets. 
The consistency of the gap across four subsets that vary
independently in cancer type and question category
constitutes the reliable signal. This result closes the evidence chain. The overlap audit
(\S\ref{sec:overlap_audit}) established that
contamination is pervasive. The feature-space analysis
(\S\ref{sec:feature_evidence}) established that it
produces detectable memorization across two structurally
distinct encoders. The meta-analysis
(\S\ref{sec:meta_analysis}) established that the highest
published accuracies co-occur with the most severe
contamination. The within-benchmark stratification
presented here isolates the effect. On the same
benchmark, the same model, the same cancer types, and the
same question categories, patients seen during training
score $3$ to $8$ percentage points higher than unseen
patients.
\section{Discussion}
\label{sec:discussion}
 
\subsection{Objections Considered}
\label{sec:objections}
 
The first objection holds that strong performance on a TCGA-overlapping test set
still demonstrates clinical competence, since TCGA broadly samples the global
oncology distribution. This misidentifies the deployment barrier. The challenge
in clinical translation is not diagnosing familiar presentations within a known
laboratory but maintaining reliability under the domain shifts that deployment
guarantees, including new scanner hardware, altered staining protocols, and
unseen patient demographics~\cite{de2025current, roberts2021common}. A
model that retrieves memorized batch artifacts rather than reasoning over
morphology will fail silently at a new institution, the failure mode that
site-specific shortcuts produce across medical
imaging~\cite{degrave2021ai, howard2021impact}. Evaluation on strictly
held-out cohorts is therefore not artificially harsh but the minimum condition
under which a generalization claim can be falsified. A second objection holds that exhaustive deduplication across millions of
multimodal pairs is computationally impractical and that some leakage is an
acceptable price of scale. This conflates two distinct operations.
Deduplicating the training datasets is expensive, but auditing the train-test
boundary is cheap and non-negotiable, and as Section~\ref{sec:experiments}
shows it is decidable in closed form from canonical TCGA identifiers
(Eq.~\ref{eq:overlap}). The premise also fails for pathology in particular. In
natural-image domains scaling absorbs diverse visual concepts, whereas
uncurated scaling in pathology typically ingests near-redundant tiles of the
same phenotype from the same patient and site, so scale amplifies the batch
confounders we document rather than diluting them. If the benchmark itself is
contaminated, no amount of training-set scaling can be verified as progress.
 
\subsection{Toward Contamination-Free Evaluation}
\label{sec:recommendations}
 
We outline three mutually reinforcing directions that together form a protocol
for credible evaluation:
 
The first lever is benchmark construction. A credible replacement benchmark
must draw its test cohort independently of every public instruction-tuning
corpus. Because TCGA is the largest public repository meeting the joint
requirement of diagnostic-grade WSIs, free redistribution, and paired reports,
removing institutional overlap requires evaluation data from Tissue Source
Sites absent from all major training datasets, which TCGA-derived sets
structurally cannot provide. We therefore recommend independently curated
multi-center holdout cohorts partitioned under preserved-site cross-validation,
so that no Tissue Source Site is split across the train-test
boundary~\cite{howard2021impact}. Each benchmark should publish its canonical
slide, case, and TSS identifiers, retain a privately held evaluation portion
that is never released, and migrate over time to submission-based
infrastructure in which models are uploaded to an isolated server rather than
test data distributed to authors~\cite{xu2022codabench}. A public and private
split with rate-limited submissions, the standard defense against leaderboard
overfitting~\cite{xu2022codabench, xu2024benchmark}, blunts the slow
test-set memorization that even held-out benchmarks otherwise suffer.
 
The second lever is provenance disclosure. Contamination is invisible today
because papers report a single headline zero-shot accuracy without disclosing
what the model has seen. Following the documentation discipline of datasheets
for datasets and the Data Provenance
Initiative~\cite{gebru2021datasheets}, we propose three
mandatory disclosures for any zero-shot generalization claim: the complete list
of training, instruction-tuning, and visual-encoder pretraining corpora
consumed by the model; the case-level and TSS-level overlap,
$\rho_{\text{case}}$ and $\rho_{\text{TSS}}$, between every disclosed dataset and
the benchmark, computed with the released audit pipeline, and stratified
accuracy on audit-clean and audit-contaminated subsets whenever overlap is
non-zero. Emitting these fields as machine-readable metadata~\cite{akhtar2024croissant} would let reviewers verify
provenance automatically rather than by manual cross-referencing.
 
The third lever is automated overlap auditing. Sustained decontamination
requires community tooling rather than per-paper diligence. Identifier-level
matching, the basis of our audit, is exact but presumes intact metadata, and
should be complemented by content-level detection that survives re-export and
relabeling, such as perceptual hashing and digital fingerprinting of
whole-slide images~\cite{jakhar2025effective}, paralleling the n-gram and
canary-based contamination detection now standard in language-model
evaluation~\cite{xu2024benchmark, deng2024investigating}. For upstream
encoder corpora that identifier matching cannot reach, the linear-probing
strategy offers a black-box test: recoverability of patient or TSS identity from frozen features is itself a
contamination signal even when the pretraining data is proprietary. Released as
a continuously maintained open pipeline against which any new corpus or
checkpoint can be checked, such tooling would convert contamination auditing
from a one-off exercise into a standing community resource.

\subsection{Limitations}
\label{sec:limitations}

Three scope conditions qualify our findings. First, the audit covers only the instruction-tuning stage, where structured TCGA metadata makes $\rho_{\text{case}}$ and $\rho_{\text{TSS}}$ computable in closed form; leakage propagated through the proprietary visual-encoder pretraining datasets, which are generally undisclosed, remains unquantified, though the linear probe offers a black-box signal for this upstream channel. Second, the within-benchmark stratification is computed on a single contaminated checkpoint by partitioning its test set into audit-clean and audit-contaminated slides within the same benchmark; because the audit-clean partition is comparatively small, the +3 to +8 pp gap is directionally consistent across four independently varying subsets but imprecisely estimated in magnitude. Third, our audit detects overlap via TCGA identifiers, so it assumes intact provenance metadata and yields a lower bound on true contamination: slides whose barcodes have been stripped, re-exported, or relabeled escape identifier matching and would require the content-level detection we recommend in Section~\ref{sec:recommendations}. Despite these conditions, our core result remains robust across two distinct encoders via exact identifier matching. These boundaries simply mark where quantification ends and our proposed replacement benchmark begins.
\section{Conclusion}
\label{sec:conclusion}
 
We present the first systematic audit of train-test contamination across the public WSI vision-language ecosystem. Tracing TCGA case, slide, and TSS barcodes across the instruction-tuning datasets and TCGA-derived benchmarks, we document 92.3\% to 100\% case-level overlap. We show that both leakage levels are linearly decodable from the feature spaces of two structurally distinct foundation encoders. Notably, the natively slide-level encoder concentrates rather than dilutes patient-identifying signals, translating into a consistent accuracy gap between leaked and clean slides. Consequently, the highest reported state-of-the-art accuracies across WSI VLMs cluster on the most heavily contaminated benchmarks. These findings do not diminish the engineering achievements of recent pathology VLMs, but establish that current zero-shot generalization claims are unreliable: present benchmarks cannot distinguish genuine multimodal reasoning from nearest-neighbor retrieval over memorized institutional and patient-specific artifacts. The remedy is a strict change in evaluation practice. Until the community enforces slide-, patient-, and TSS-level deduplication, discloses cohort overlaps, and evaluates on independently curated multi-center holdouts, reported gains in WSI vision-language modeling cannot be separated from memorization artifacts.


\bibliographystyle{splncs04}
\bibliography{sample}

\end{document}